\documentclass[runningheads]{llncs}
\usepackage[T1]{fontenc}
\usepackage[USenglish]{babel} 
\usepackage[ansinew]{inputenc}
\usepackage[noadjust]{cite}
\usepackage{lmodern} 
\usepackage{graphicx} 
\usepackage{float}
\usepackage{textcomp}
\usepackage{amsmath}
\usepackage{amsfonts}
\usepackage{amssymb}
\usepackage{bm}
\usepackage{url}
\usepackage{hyperref}
\hypersetup{ backref=true,       
		pagebackref=true,               
            hyperindex=true,                
            colorlinks=true,                
            breaklinks=true,                
            urlcolor= blue,                
            linkcolor= black,
		citecolor=green
}
\usepackage{booktabs}
\usepackage{subcaption}
\usepackage{tabularx}
\usepackage{makecell}
\usepackage{romannum}
\usepackage[super]{nth}
\usepackage[dvipsnames]{xcolor}
\usepackage{academicons}
\usepackage{dsfont}
\usepackage{flafter}
\usepackage{multirow}
\newcommand{\vecc}[1]{\mathbf{#1}}

\newcommand{\tpx}{\text{TP}_{\text{x}}}
\newcommand{\fpx}{\text{FP}_{\text{x}}}
\newcommand{\fnx}{\text{FN}_{\text{x}}}

\newcommand{\Dex}{D_{\text{bnds}}}

\newcommand{\Nft}{\text{N}_{\text{ft}}}
\newcommand{\Nl}{\text{N}_{\text{l}}}

\newcommand{\CDT}{\text{DT}}
\newcommand{\Pcls}{\text{P}_{\text{cls}}}
\newcommand{\Pcat}{\text{P}_{\text{cat}}}
\newcommand{\Psdc}{\text{P}_{\text{sdc}}}
\newcommand{\Rcls}{\text{R}_{\text{cls}}}
\newcommand{\Rcat}{\text{R}_{\text{cat}}}
\newcommand{\Rsdc}{\text{R}_{\text{sdc}}}
\newcommand{\Px}{\text{P}_{\text{x}}}
\newcommand{\Rx}{\text{R}_{\text{x}}}

\newcolumntype{L}[1]{>{\raggedright\let\newline\\\arraybackslash\hspace{0pt}}m{#1}}
\newcolumntype{C}[1]{>{\centering\let\newline\\\arraybackslash\hspace{0pt}}m{#1}}
\newcolumntype{R}[1]{>{\raggedleft\let\newline\\\arraybackslash\hspace{0pt}}m{#1}}

\usepackage{pgfplots, gincltex}
\usepackage{lipsum}
\usepackage{tikz}
\usepackage{wrapfig}
\usetikzlibrary{patterns}
\pgfplotsset{compat=newest}
\pgfplotsset{plot coordinates/math parser=false}
\usepackage{placeins}
\usepackage{array}

\begin{document}

\title{\bf A Low-cost Strategic Monitoring Approach for Scalable and Interpretable Error Detection in Deep Neural Networks}
\titlerunning{Monitoring Deep Neural Networks}

\author{Florian Geissler\inst{1} \and Syed Qutub\inst{1} \and Michael Paulitsch\inst{1} \and Karthik Pattabiraman\inst{2}}
\institute{Intel Labs, Germany \\
\email{\{florian.geissler, michael.paulitsch, qutub.syed\}@intel.com}\\
\and University of British Columbia, Canada\\
\email{karthikp@ece.ubc.ca}}
\authorrunning{Florian Geissler, Syed Qutub, Michael Paulitsch, and Karthik Pattabiraman}
\date{\today}
\maketitle

\begin{abstract}
We present a highly compact run-time monitoring approach for deep computer vision networks that extracts selected knowledge from only a few (down to merely two) hidden layers, yet can efficiently detect silent data corruption originating from both hardware memory and input faults. 
Building on the insight that critical faults typically manifest as peak or bulk shifts in the activation distribution of the affected network layers, we use strategically placed quantile markers to make accurate estimates about the anomaly of the current inference as a whole.
Importantly, the detector component itself is kept algorithmically transparent to render the categorization of regular and abnormal behavior interpretable to a human.
Our technique achieves up to ${\sim}96\%$ precision and ${\sim}98\%$ recall of detection.
Compared to state-of-the-art anomaly detection techniques, this approach requires minimal compute overhead (as little as $0.3\%$ with respect to non-supervised inference time) and contributes to the explainability of the model.
\end{abstract}

\section{Introduction}

\begin{figure}[tbp]
\centering
\includegraphics[width=0.7\textwidth]{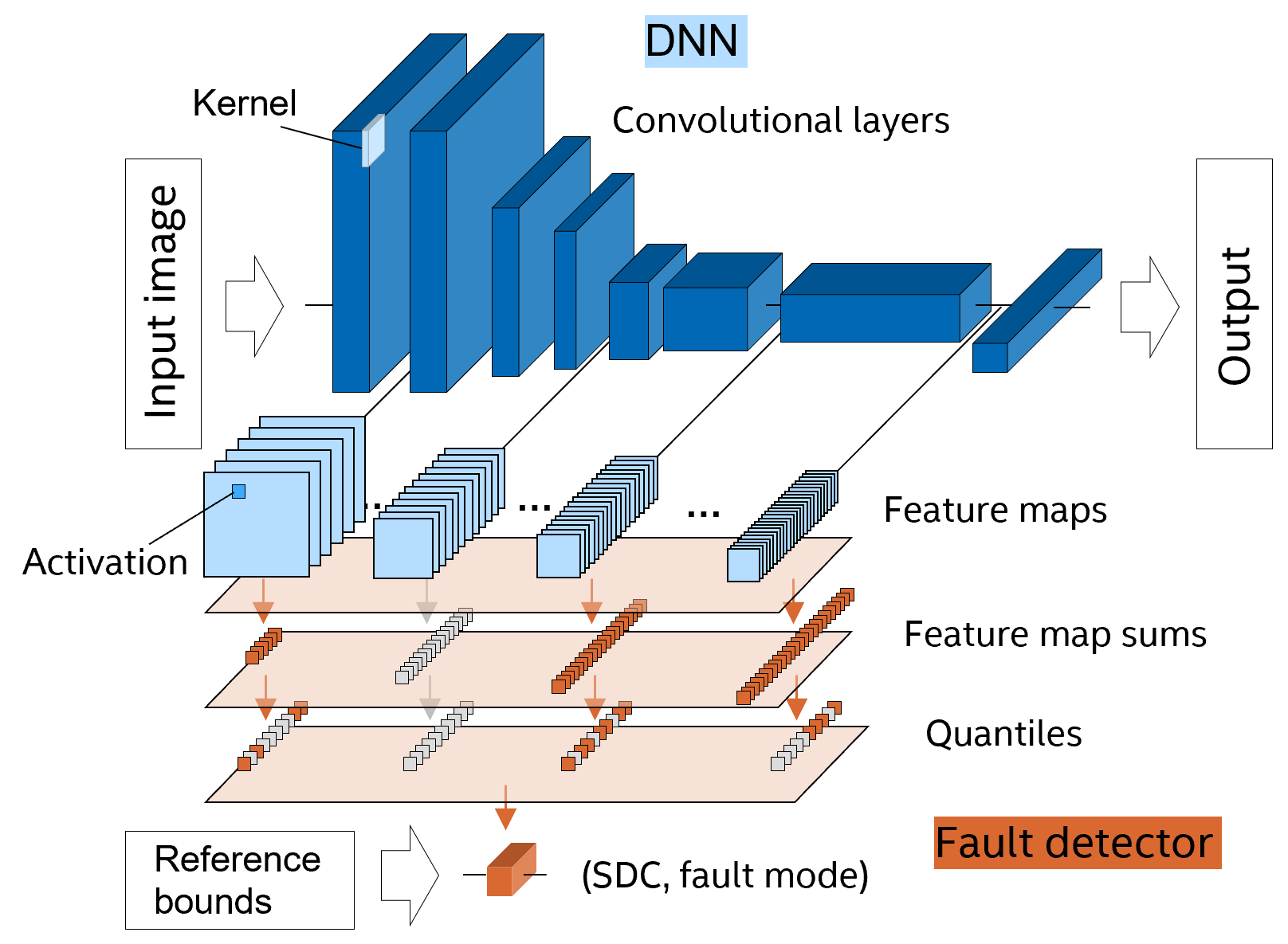}
\setlength{\belowcaptionskip}{-10pt}
\caption{Monitoring architecture for quantile shift detection.}
\label{fig:sys}
\end{figure}

Deep neural networks (DNNs) have reached impressive performance in computer vision problems such as object detection, making them a natural choice for problems like automated driving \cite{Balasubramaniam2022}. However, DNNs are known to be highly vulnerable to faults. For example, even small changes to the input such as adding a customized noise pattern that remains invisible to the human eye, can stimulate silent prediction errors \cite{Goodfellow2015}.
Similarly, modifying a single out of millions of network parameters, in the form of a bit flip, is sufficient to cause severe accuracy drops \cite{Hong2019}. 

Because DNNs are being deployed in safety-critical applications such as autonomous vehicles (AVs), we need efficient mechanisms to detect errors that cause such silent data corruptions (SDC). 
Beyond the functional part, trust in the safety of the application requires that the error detectors are interpretable by the user, so that he/she can develop an intuitive understanding of the regular and irregular behavior of the network \cite{BarredoArrieta2019}. In an AV, for example, a user who does not trust an automated perception component due to its opaque decision-making, will not trust a black-box fault monitor either. Therefore, it is important to build interpretable error detectors for DNNs.

The goal of error detection is to supervise a small, yet representative subset of activations - during a given network inference - for comparison with a previously extracted fault-free baseline. This leads to three key challenges: \textbf{(1)} How can one compress the relevant information into efficient abstractions?
\textbf{(2)} How can one efficiently perform the anomaly detection process, for complex patterns? \textbf{(3)} Can the anomaly detection decision be understandable to a human, so that insights are gained about the inner workings of the network?

\begin{figure}[tbp]
        \centering
	\begin{subfigure}{1.\textwidth}
	\includegraphics[width=\textwidth]{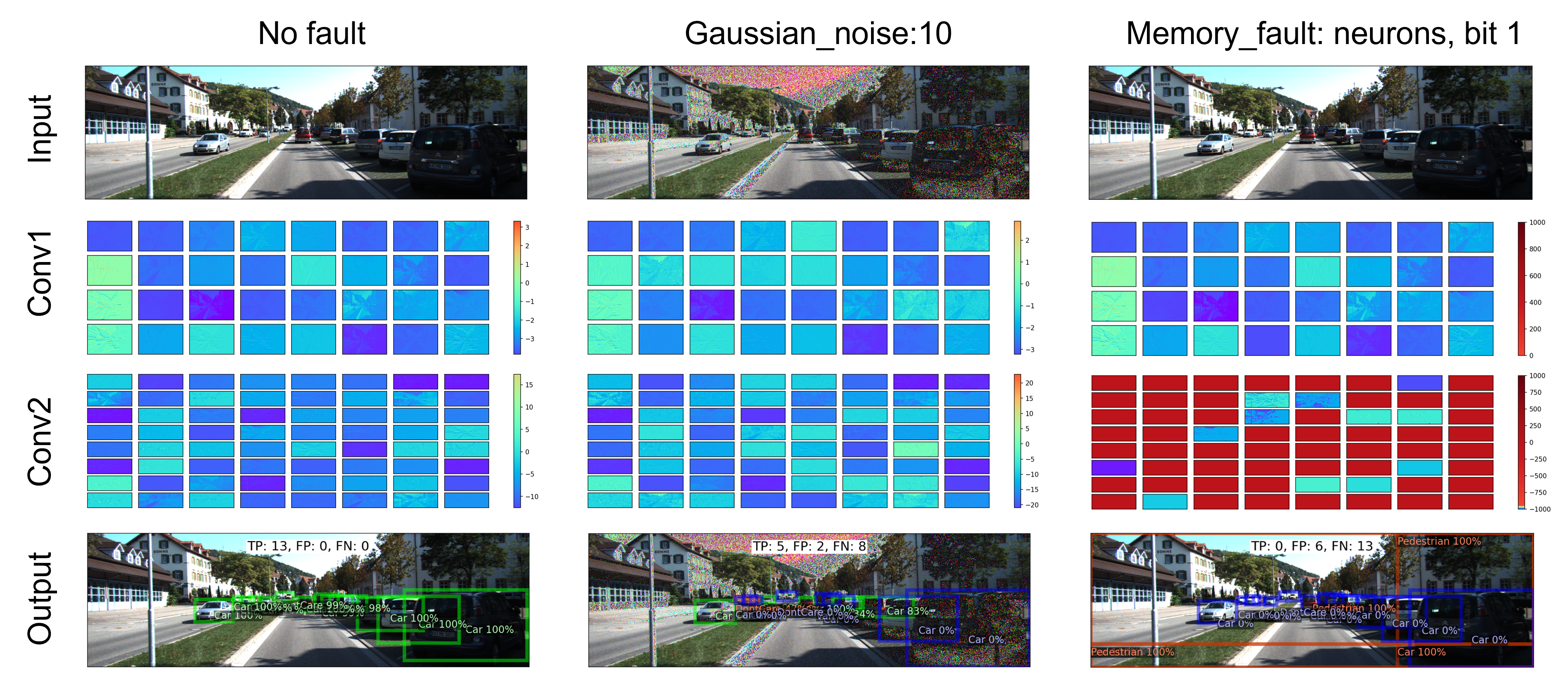}
 \setlength{\belowcaptionskip}{-10pt}
	\caption{Feature map visualization, the color code represents activation magnitudes. The final prediction result is given as inset in the output row (TP: True positives, FP: false positives, FN: false negatives).}
	\label{fig:ex0} 
	\end{subfigure}
\bigbreak
	\begin{subfigure}{0.499\textwidth}
	\centering
	\includegraphics[width=0.81\textwidth]{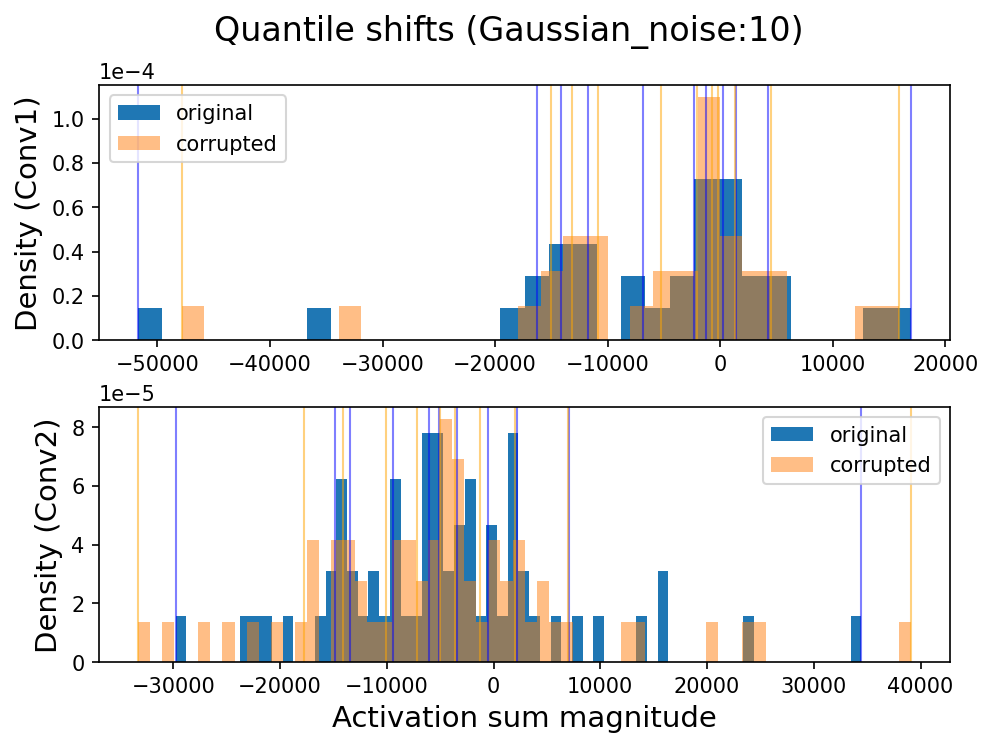}
	\caption{Original image vs. noise corruption of (a). We show both activation sums (bars) and quantile values (overlayed vertical lines).}
	\label{fig:ex1}
	\end{subfigure}
	\hspace{0.05cm}
	\begin{subfigure}{0.4812\textwidth}
	\centering
	\includegraphics[width=0.81\textwidth]{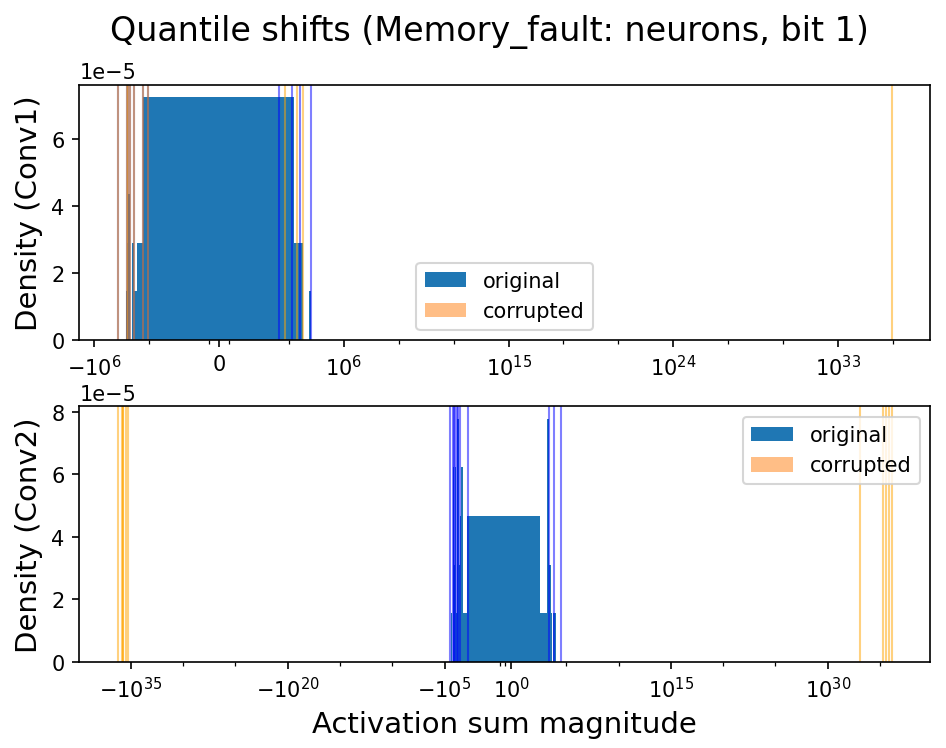}
	\caption{Original image vs. memory fault of (a) (symmetric log scale). Large corruption magnitude ranges lead to histogram densities ${<}10^{-30}$.
 }
	\label{fig:ex2}
	\end{subfigure} \\
\setlength{\belowcaptionskip}{-10pt}
\caption{
\textbf{(a)} The feature map appearance is slightly changed with noise and massively affected by the memory FI.
\textbf{(b)} Noise causes a small shift of multiple quantiles from the affected layer onwards (activation bulk shift). 
\textbf{(c)} The layer with the memory FI shows a large shift of the maximum quantile (activation peak shift), which then propagates to other quantiles. 
}
\label{fig:ex} 
\end{figure}

Unfortunately, no existing approach satisfactorily addresses all three of the above challenges (Sec.~\ref{sec:relatedwork}).
This paper presents an solution using a monitoring architecture that taps into the intermediate activations only at selected strategic points and interprets those signatures in a transparent way, see Fig.~\ref{fig:sys}. 
Our approach is designed to detect SDC-causing errors due to input corruptions \textit{or} hardware faults in the underlying platform memory. Our main observation that underpins the method is that an SDC occurs when a fault {\em either} increases the values of a few activations by a large margin (referred to here as an \textit{activation peak shift}), or the values of many activations each by a small margin (\textit{activation bulk shift}). As  Fig.~\ref{fig:ex} shows, the former is observed typically for platform faults, while the latter is observed for input faults.
We then use discrete quantile markers to distill the knowledge about the variation of the activation distribution in a given layer. 
Conceptually, within a faulty layer, we can expect a large change of only the top quantiles for activation peak shifts, and small changes of the lower and medium quantiles for bulk shifts (Fig.~\ref{fig:ex}). This idea allows us to produce discriminative features for anomaly detection from a small number of monitored elements, with a single detector.

In summary, we make the following contributions in this paper: 
\begin{itemize}
\item We demonstrate that even for complex object detection networks, we can identify anomalous behavior from quantile shifts in only a few layers.
\item We identify minimal sets of relevant features and discuss their universality across models.
\item We efficiently differentiate input and hardware fault classes with a single detector.
\item We show that the anomaly detection process can be achieved with algorithmically transparent components, such as decision trees.
\end{itemize}

The article is structured as follows: Sec.~\ref{sec:relatedwork} discusses related work, while Sec.~\ref{sec:setup} describes our experimental setup. 
We present our method in Sec.~\ref{sec:model}, and the results of our evaluation in Sec.~\ref{sec:results}. 

\section{Related Work}
\label{sec:relatedwork}
There are three main categories of related work. 

\textbf{Image-level Techniques}: 
Input faults can be detected from the image itself (i.e., before network inference), in comparison with known fault-free data, resulting for example in specialized blur detectors \cite{Huang2018}. 
However, these techniques do not necessarily relate to SDC in the network, as image-level corruptions may be tolerated by the model.

\textbf{Activation Patterns}:
Methods to extract activation patterns range from activation vectors \cite{Cheng2019} to feature traces \cite{Schorn2018, Schorn2020}. 
However, these techniques do not scale well to deeper models as they result in a massive number of monitored features and large overheads.
Zhao et al. \cite{Zhao2022} attempt to reduce the monitoring effort by leveraging only activations from selected layers and compressing them with customized convolution and pooling operations. 
This leads to a rather complex, non-interpretable detector component, and the selection of monitored layers remains empirical. 

\textbf{Anomaly Detection} techniques establish clusters of regular and anomalous data to efficiently categorize new input. 
In single-label problems, such as image classification, fault-free clusters are typically formed by samples that belong to the same individual label \cite{Henzinger2020}, suggesting that those samples also share common attributes in the space of intermediate activations. 
This technique does not generalize to multi-label problems though, such as object detection, as many objects (in the form of bounding boxes and labels) are represented in the same image.  
More abstracted clustering rules such as the maximum activation range per layer have been proposed~\cite{Li2017, Chen2021}. However, these detectors omit more subtle errors within the activation spectrum, for example resulting from input faults.
In other work~\cite{Schorn2018, Schorn2020, Zhao2022}, a secondary neural network is trained to perform the detection process. 
This comes at the cost that the detector then does not feature algorithmic transparency~\cite{BarredoArrieta2019} and hence the anomaly decision is not understandable to a human.
The same limitations are found in the context of detector subnetworks that are trained to identify adversarial perturbations \cite{Metzen2017}.

\textbf{Summary}: We see that none of the prior techniques satisfactorily address the challenges outlined earlier. We present a new technique to overcome this problem in this paper.

\section{Experimental Setup and Preliminary Study}
\label{sec:setup}

\textbf{Models and Datasets:}
We use the three classic object detection networks Yolo(v3), Single Shot Detector (SSD), and RetinaNet from the \textit{open-mmlab} \cite{mmdetection} framework, as well as the two standard image classification networks ResNet50 and AlexNet from \textit{torchvision} \cite{Paszke2019}. Object detection networks are pretrained on Coco \cite{Microsoft2017} and were retrained on Kitti \cite{Geiger2013}, with the following AP50 baseline performances: Yolo+Coco: $55.5\%$, Yolo+Kitti: $72.8\%$, SSD+Coco: $52.5\%$, SSD+Kitti: $66.5\%$, RetinaNet+Coco: $59.0\%$, RetinaNet+Kitti: $81.8\%$.  
Image classification models were pretrained on ImageNet \cite{JiaDeng2009}, providing accuracies of $78.0\%$ (ResNet) and $60.0\%$ (AlexNet) for the test setup. 
The data was split in a ratio of $2{:}1$ for detector training and testing.
All models are run in \textit{Pytorch} with the IEEE-standard FP32 data precision \cite{IEEE2019}.

\textbf{Fault Modes:}
Input faults are modeled using \textit{torchvision} \cite{Paszke2019} transform functions and are applied in three different magnitudes to the raw RGB images. 
We select three perturbation patterns that are popular in computer vision benchmarks such as ImageNet-C \cite{Hendrycks2019} for our analysis: i) \textit{Gaussian noise} due to low lighting conditions or noise in the electronic signal of the sensor device. Low ($0.1$), medium ($1$), and high ($10$) noise is tested. ii) \textit{Gaussian blur}, reflecting for example a camera lens being out of focus. We choose a kernel size of $(5,9)$ and a symmetric, variable standard deviation ($0.3,1,3$). iii) \textit{Contrast reductions} simulate poor lighting conditions or foggy weather. We adjust the contrast by a factor between zero (no contrast, gray image) and one (original image).
The selected models have different vulnerabilities to input faults, for example, the two image classification models ResNet and AlexNet are highly sensitive to contrast adjustments, but are rather robust to noise and blur faults. For the remaining models, the trend is reversed. 

Hardware faults are modeled as single bit flips in the underlying memory and injected using \textit{PytorchAlfi} \cite{Graefe2022}. 
Such flips can occur randomly either in the buffers holding temporary activation values (\textit{neuron} fault), or in dedicated memory which holds the parameters of the network (\textit{weight} faults). We group both neuron and weight faults into a single class \textit{memory} fault. This approach is in line with previous work \cite{Li2017, Schorn2018, Schorn2020, Geissler2021, Qutub2022, Hoang2020, Chen2021}. We target all convolutional layers.

\textbf{Fault Metrics:}
First, detectable uncorrectable errors (DUE) can occur when invalid symbols such as \textit{NaN} or \textit{Inf} are found among the activations at inference time.
During fault injection, we observe DUE events only for memory faults, with rates ${<}1 \%$ across all models. DUEs can be generated also at the detector stage, in the process of adding up feature map sums that contain platform errors. The rates for such events vary between $0.2\%$ and $5.1\%$ with our method.
While DUE errors may affect the system's availability, they are considered less critical as they are readily detectable and there is no need for further activation monitoring \cite{Geissler2021}.

In this article, we are concerned therefore only with silent data corruption (SDC), events that lead to a silent alteration of the predicted outcome. For image classification networks, this is represented by a change in the top-1 class prediction. For object detection systems, we use an asymmetric version of the IVMOD metric \cite{Qutub2022} as SDC criterion, i.e., an image-wise increment in the FP or FN object numbers is counted as SDC.
Each experiment was done with a subset of $100$ images of the test data set, running $100$ random FIs on each image individually. For hardware faults, SDC rates are typically low (${\sim}1{-}3\%$) since drastic changes will result only from bit flips in the high exponential bits of the FP32 data type \cite{Li2017, Geissler2021}. Therefore, an additional $500$ epochs with accelerated FI only into the three highest exponential bits are performed for both flavors of memory faults. Overall, the faulty and fault-free data is found to be balanced at a ratio of about $2{:}1$.


\section{Model}
\label{sec:model}

\textbf{Notational Remarks:}
We use the range index convention, i.e., a vector is given as $\vecc{x} = (x_i)=(x^i)$, a matrix reads $\vecc{A} = (A_{ij})$, and similarly for higher-dimensional tensors.

\textbf{Monitoring Approach:}
Let us denote a four-dimensional activation tensor that represents an intermediate state of a convolutional neural network as $\vecc{T}=({T}_{n,c,h,w}) \in \mathds{R}^{N\times C\times H\times W}$,
where $N$ is the sample number, $C$ the number of channels, $H$ the height, and $W$ the width. We list $n$ as running global sample index, where samples may be further grouped in batches.
An output tensor of a specific layer $l\in[1,\ldots L]$ shall be given as $\vecc{T}^l$, with $L$ being the total number of monitored layers of the model. 
Subsets of a tensor with fixed $n,c$ are called feature maps.
Our monitoring approach first performs the summation of individual feature maps and subsequently calculates quantile values over the remaining kernels, see Fig.~\ref{fig:sys},
\begin{align}
({F}_{n,c})^l &= \sum_{h, w} ({T}_{n,c,h,w})^l, \label{eq:sum}\\
{(q_n)}_{p}^l &= \left(Q_p(({F}_{n,c})^l)_n\right). 
\label{eq:quant}
\end{align}
Here $Q_p$ is the quantile function for the percentile $p$ which acts on the $n$-th row of $({F}_{n,c})^l$. In other words, $Q_p$ reduces the kernel dimensions $c$ to a set of discrete values where we use the $10$-percentiles, i.e., $p \in [0,10,20, 30, \ldots ,90,100]$.
The result is a quantile value set, $q_p$, for a given image index $n$ and layer $l$. Note that both the summation and the quantile operations (and hence the detector) are invariant under input perturbations such as image rotations.

\textbf{Supervised Layers:} 
We intercept the output activations of all convolutional layers, as those layers provide the vast majority of operations in the selected computer vision DNNs. Yet, the same technique can be applied to any neural network layer.

\textbf{Reference Bound Extraction:}
Applied to a separate data set $\Dex$, the above technique is used pre-runtime to extract reference bounds which represent the minimum and  maximum feature sums during fault-free operation:
\begin{align}
    \begin{split}
        q_{p, {\text{min}}}^l = \min_{n \in \Dex}\left(({q}_n)_p^l\right), \\
        q_{p, {\text{max}}}^l = \max_{n \in \Dex}\left(({q}_n)_p^l\right),
    \end{split}
\label{eq:qminmax}
\end{align}

For $\Dex$, we randomly select $20\%$ of the training data \cite{Chen2021}.

\textbf{Anomaly Feature Extraction:} For a given input during runtime, Eqs.~(\ref{eq:sum}) to (\ref{eq:quant}) are used to obtain the quantile markers of the current activation distribution.
Those are further processed to a so-called \textit{anomaly feature vector} which quantifies the similarity of the observed patterns with respect to the baseline references of Eq.~\ref{eq:qminmax},
\begin{align}
{q}_p^l & \to \frac{1}{2}\left(f_{\text{norm}}(q_p^l, q_{p,{\text{min}}}^l, q_{p,{\text{max}}}^l) + 1\right).
\label{eq:qpl_final}
\end{align}
Here, $f_{\text{norm}}$ normalizes the monitored quantiles to a range of $(-1,1)$ by applying element-wise ($\epsilon=10^{-8}$ is a regularization offset)
\begin{align}
\label{eq:fnorm}
f_{\text{norm}}(a,a_{\text{min}}, a_{\text{max}}) 
  = \begin{cases}
	\tanh \left(\frac{a - a_{\text{max}}}{\left|a_{\text{max}}\right| + \epsilon} \right) & \text{if } a \geq a_{\text{min}},\\
	\tanh \left(\frac{a_{\text{min}} - a}{\left|a_{\text{min}}\right| + \epsilon} \right) & \text{if } a < a_{\text{min}}.
\end{cases}
\end{align}
Intuitively, the result of Eq.~\ref{eq:fnorm} will be positive if $a$ is outside the defined minimum ($a_{\text{min}}$) and maximum ($a_{\text{max}}$) bounds (approaching $+1$ for very large positive or negative values). The function is negative if $a$ is within the bounds (lowest when barely above the minimum), and will become zero when $a$ is of the order of the thresholds. 
In  Eq.~\ref{eq:qpl_final}, a shift brings features to a range of $(0,1)$ to facilitate the interpretation of feature importance.
%
Finally, all extracted features are unrolled into a single anomaly feature vector $\vecc{q}=((q^l)_p) = [{q}^1_{0}, q^2_{0}, \ldots q^L_{0}, q^1_{10}, \ldots q^L_{100}]$,
that will be the input to the anomaly detector component.

\textbf{Anomaly Detector:}
We use a decision tree \cite{Hastie2009} approach to train an interpretable classifier, leveraging the \textit{sklearn} package \cite{Pedregosa2011}.
The class weights are inversely proportionally to the number of samples in the respective class to compensate for imbalances in the training data.
As a measure of the split quality of a decision node we use the Gini index \cite{Hastie2009}.
To avoid overfitting of the decision tree, we perform cost-complexity pruning \cite{Pedregosa2011} with a factor varying between $1\times 10^{-5}$ and $2\times 10^{-5}$, that is optimized for the respective model.

To investigate fault class identification, we study three different detector modes with varying levels of fault class abstractions and quantify each mode $x \in \{{cls}, {cat}, {sdc}\}$ by precision, $\Px = \tpx /(\tpx + \fpx)$ and recall $\Rx = \tpx/(\tpx + \fnx)$. Here we abbreviated true positives (TP), false positives (FP), and false negatives (FN). In the class mode ($cls$), we consider only those detections as true positives where the predicted and actual fault modes (see Sec. \ref{sec:setup}) coincide exactly. Cases where SDC is detected correctly but the fault class does not match will be counted as either FP or FN in this setting. In the category mode ($cat$), those SDC detections are considered true positives where the predicted and actual fault class fall into the same category of either \textit{memory fault} or $\textit{input fault} = \{noise, blur, contrast\}$. That means, fault class confusions within a category will not reduce the performance in this mode. 
The final precision and recall values for the class and category mode are given as the average over all classes or categories, respectively. Finally, in the mode $sdc$, we consider all cases as true positives where SDC was correctly identified regardless of the specific fault class. This reflects a situation where one is only interested in the presence of SDC overall, rather than the specific fault class.

\section{Results}
\label{sec:results}

\begin{table}[tbp]
\caption{Precision ($P$), Recall ($R$), and decision tree (DT) complexity - given as the number of used features ($\Nft$) and monitored layers ($\Nl$) - for different setups. 
Every detector was retrained $10$ times with different random seeds and the averages across all runs are given. Errors are shown when relevant to the given rounding.
We list both the classifiers making use of all extracted quantiles (\textit{full}) and the averaged reduced (\textit{red}) detector models, where guided feature reduction was applied, see Fig.~\ref{fig:pr}. 
Best-in-class detectors are highlighted in each column.
}
\vspace{.15cm}
\centering
	\begin{tabular}{C{3.5cm}|C{1cm}C{1cm}C{1cm}|C{1cm}C{1cm}C{1cm}|C{1.5cm}|}
	{\textbf{Model}} & \multicolumn{3}{c|}{\textbf{P}($\%$)} &  \multicolumn{3}{c|}{\textbf{R}($\%$)} & \multicolumn{1}{c|}{\textbf{$\CDT$}}  \\ \cline{2-8}
	 & $\Pcls$ & $\Pcat$ & $\Psdc$ & $\Rcls$ & $\Rcat$ & $\Rsdc$ &  $\Nft/\Nl$ \\ \hline \hline
	\textbf{Yolo+Coco} &  &  &  &  & & &  \\ 
	full & $95.8$ & $96.4$ & $96.1$ & $98.2$ & $98.6$ & $98.4$ & $825/75$  \\
	red (avg) & $93.3$ & $94.6$ & $93.4$ & $97.4$ & $96.3$ & $96.7$ & $2/2$  \\ \hline
	\textbf{Yolo+Kitti} &  &  &  &  & & &  \\ 
	full & $97.3$ & $97.5$ & $97.4$ & $\bf{99.1}$ & $99.3$ & $99.2$ & $825/75$   \\
	red (avg) & $92.6$ & $92.1$ & $92.0$ & $97.3$ & $96.4$ & $96.8$ & $3/2$   \\ \hline
	\textbf{SSD+Coco} &  &  &  &  & & &  \\ 
	full & $96.6$ & $97.2$ & $96.6$ & $98.2$ & $98.5$ & $98.3$ & $429/39$   \\
	red (avg) & $95.2$ & $96.3$ & $94.9$ & $96.5$ & $94.5$ & $95.9$ & $3/3$   \\ \hline
	\textbf{SSD+Kitti} &  &  &  &  & & &  \\ 
	full & $96.0$ & $97.1$ & $96.2$ & $98.4$ & $98.7$ & $98.6$ & $429/39$  \\
	red (avg) & $92.8$ & $94.6$ & $92.1$ & $98.0$ & $97.7$ & $98.2$ & $2/2$  \\ \hline
	\textbf{RetinaNet+Coco} &  &  &  &  & & & \\ 
	full & $96.6$ & $95.7$ & $96.9$ & $97.1$ & $94.9$ & $98.0$ & $781/71$   \\
	red (avg) & $\bf{96.6}$ & $96.6$ & $96.5$ & $97.0$ & $94.6$ & $98.2$ & $2/2$  \\ \hline
	%
	\textbf{RetinaNet+Kitti} &  &  &  &  & & &  \\ 
	full & $\bf{97.5}$ & $97.3$ & $97.5$ & $98.6$ & $98.2$ & $98.7$ & $781/71$   \\
	red (avg) & $96.2$ & $96.6$ & $95.9$ & $\bf{98.6}$ & $97.8$ & $98.9$ & $2/2$   \\ \hline
	\textbf{ResNet+Imagenet} &  &  &  &  & & &  \\ 
	full & $93.9$ & $\bf{98.3}$ & $\bf{97.6}$ & $98.1$ & $\bf{99.6}$ & $\bf{99.4}$ & $583/53$ \\
	red (avg) & $92.1$ & $\bf{97.6}$ & $\bf{96.7}$ & $98.3$ & $\bf{99.6}$ & $\bf{99.5}$ & $3/3$  \\  \hline
	\textbf{AlexNet+Imagenet} &  &  &  &  & & & \\ 
	full & $96.1$ & $\bf{98.3}$ & $97.3$ & $98.4$ & $99.2$ & $99.0$ & $55/5$  \\ 
	red (avg) & $93.2$ & $96.8$ & $95.0$ & $98.0$ & $99.0$ & $98.8$ & $4/3$  \\  \hline
	\end{tabular}
	\label{tab:res1}
\end{table}

\subsection{Detector Performance}
\label{sec:featurereduction}

\textbf{Error Detection:}
Tab.~\ref{tab:res1} shows the precision, recall, and decision tree complexity for the studied detectors and models. 
When all extracted features are leveraged by the decision tree classifier (referred to as \textit{full} model), the average class-wise detection precision varies between $93.9\%$ (ResNet) and $97.5\%$ (RetinaNet+Kitti),  while the recall is between $97.1\%$ (RetinaNet+Coco) and $99.1\%$ (Yolo+Kitti). 
If only the fault category needs to be detected correctly, we find $\Pcat>95\%$ and $\Rcat>94\%$. Correct decisions about the presence of SDC only are done with $\Psdc>96\%$ and $\Rsdc\geq98\%$.
Across models, we observe (not shown in Tab.~\ref{tab:res1}) that the most common confusion are false positive noise detections, leading to a reduced precision in the individual \textit{noise} class (worst case $75.8\%$ for ResNet).
The recall is most affected by memory faults (lowest individual class recall $90.6\%$ for RetinaNet+Coco).

\begin{figure}[tbp]
\includegraphics[width=.48\textwidth]{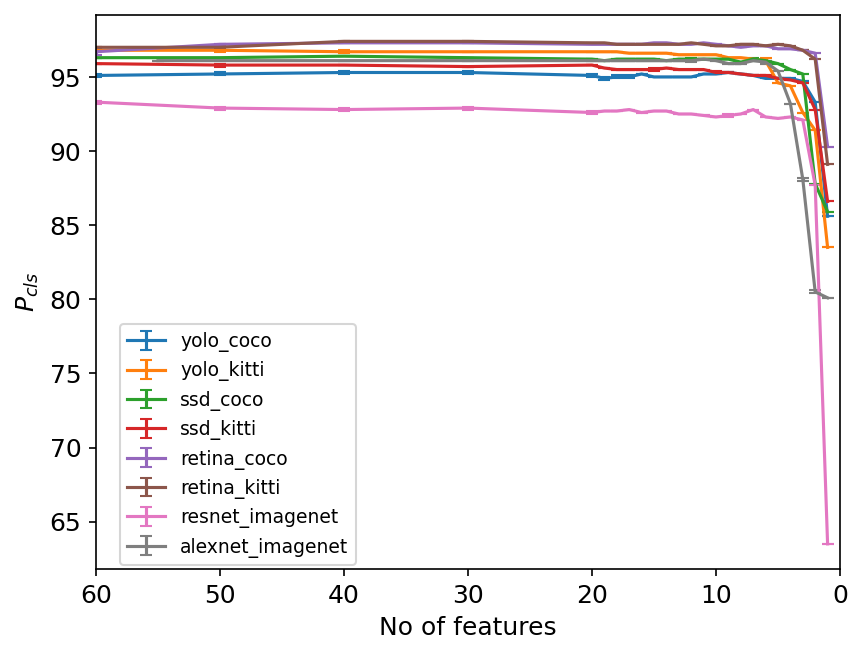}
\includegraphics[width=.48\textwidth]{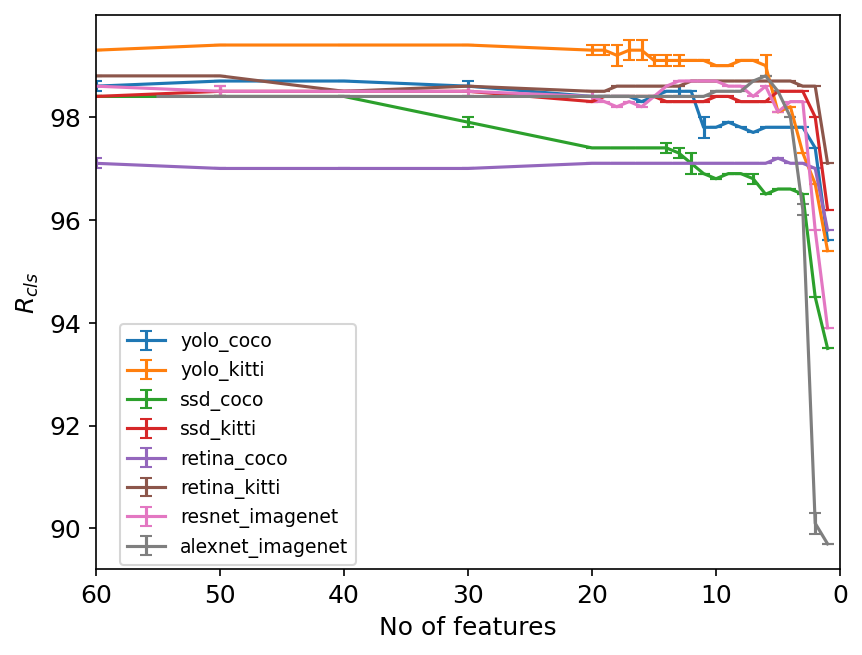}
\setlength{\belowcaptionskip}{-10pt}
\caption{Precision and recall of class-wise SDC detection when reducing the number of monitored features (average of $10$ independent runs).}
\label{fig:pr}
\end{figure}

The detection rates of the full model in Tab.~\ref{tab:res1} outperform the ones reported in the comparable approach of Schorn et al. \cite{Schorn2018} (using feature map tracing) and the blur detection in Huang et al. \cite{Huang2018} in terms of precision and recall. When using alternative metrics (not shown in Tab.~\ref{tab:res1}) for comparison with other detector designs, we find that our method achieves class-wise misclassification rates ranging between $0.7\%$ and $2.0\%$, depending on the model, which is on par with the results for example in Cheng et al.~\cite{Cheng2019}. Similarly, the calculated class-wise true negative rates vary between $99.6\%$ and $99.8\%$, reaching or exceeding the classifier performance in Zhao et al.~\cite{Zhao2022}.
Note that all mentioned references are limited to image classification networks.

\textbf{Feature Reduction:}
The number of monitored features can be drastically reduced without significantly affecting the detection performance. This means that many quantiles represent similar information and further distillation can be applied. For feature reduction, we follow two steps: 
First, all quantile features of the full model are ranked according to their Gini importance \cite{Pedregosa2011} in the decision tree. Then, we retrain the classifier with a successive number of features, starting from the most important one only, to the two most important ones, etc. 
A reduced model is accepted as efficient if it recovers at least $95\%$ of both the precision and recall performance of the original model with all features.

Fig.~\ref{fig:pr} shows the results of the feature reduction. Inspecting performance trends from larger to smaller feature numbers, we observe that the detection rate stagnates over most of the elimination process, before dropping abruptly when the number of used features reduces beyond a limit. 
On average, the number of monitored features and layers that are required to maintain close-to-original performance (as defined above) are as few as $2$ to $4$ and $2$ to $3$, respectively. 
For a model like Yolo, this means that only $2$ out of the $75$ convolution layers have to be supervised.
The average characteristics of the resulting detector models is shown in Tab.~\ref{tab:res1} as reduced (\textit{red}) model.
%

\subsection{Minimal Monitoring Features}
\label{sec:minmodel}
\textbf{Minimal Feature Search:}
The feature reduction process in Sec.~\ref{sec:featurereduction} demonstrates that only few strategic monitoring markers are needed to construct an efficient detector model. In this section, we elaborate further to what extent the model can be compressed, and which features are the most relevant. We apply the following strategy, starting from a full classifier model using all quantile features: 
1) Apply the feature reduction technique described in Sec.~\ref{sec:featurereduction} to identify minimal monitoring features that maintain at least $95\%$ of the original precision and recall. This combination of features is added to a pool of minimal model candidates. 2) A new instance of the full model is initiated and all feature candidates from the pool are eliminated. Return to the first step to find alternative candidates until a certain search depth (we choose $24$) is exhausted.

\begin{figure}[tbp]
\centering
	\begin{subfigure}{0.45\textwidth}
	(a) Yolo\\
	\hspace*{-0.3cm}\includegraphics[width=1.05\textwidth]{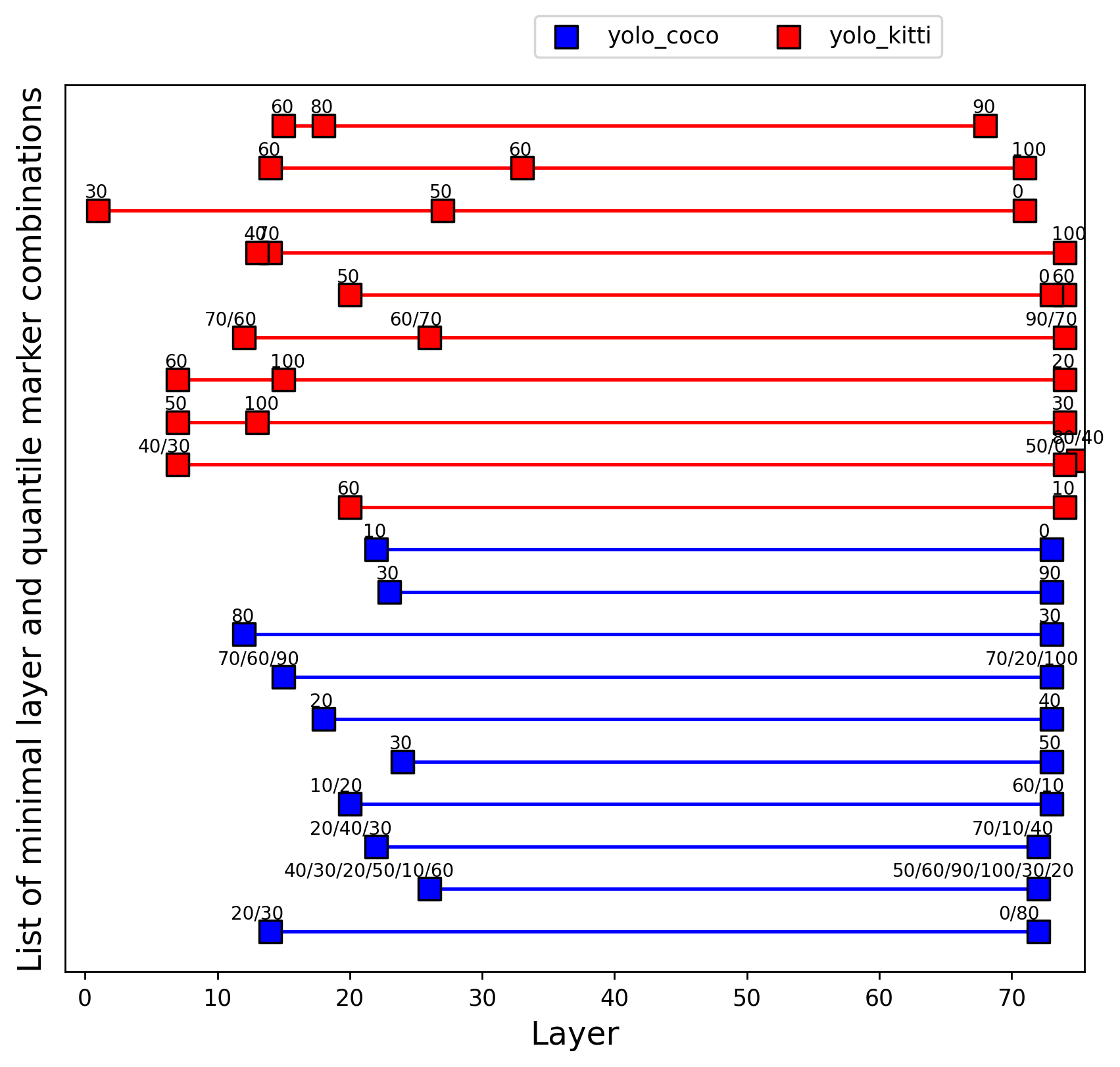}
	\end{subfigure}
	\begin{subfigure}{0.45\textwidth}
	(b) SSD\\
	\includegraphics[width=\textwidth]{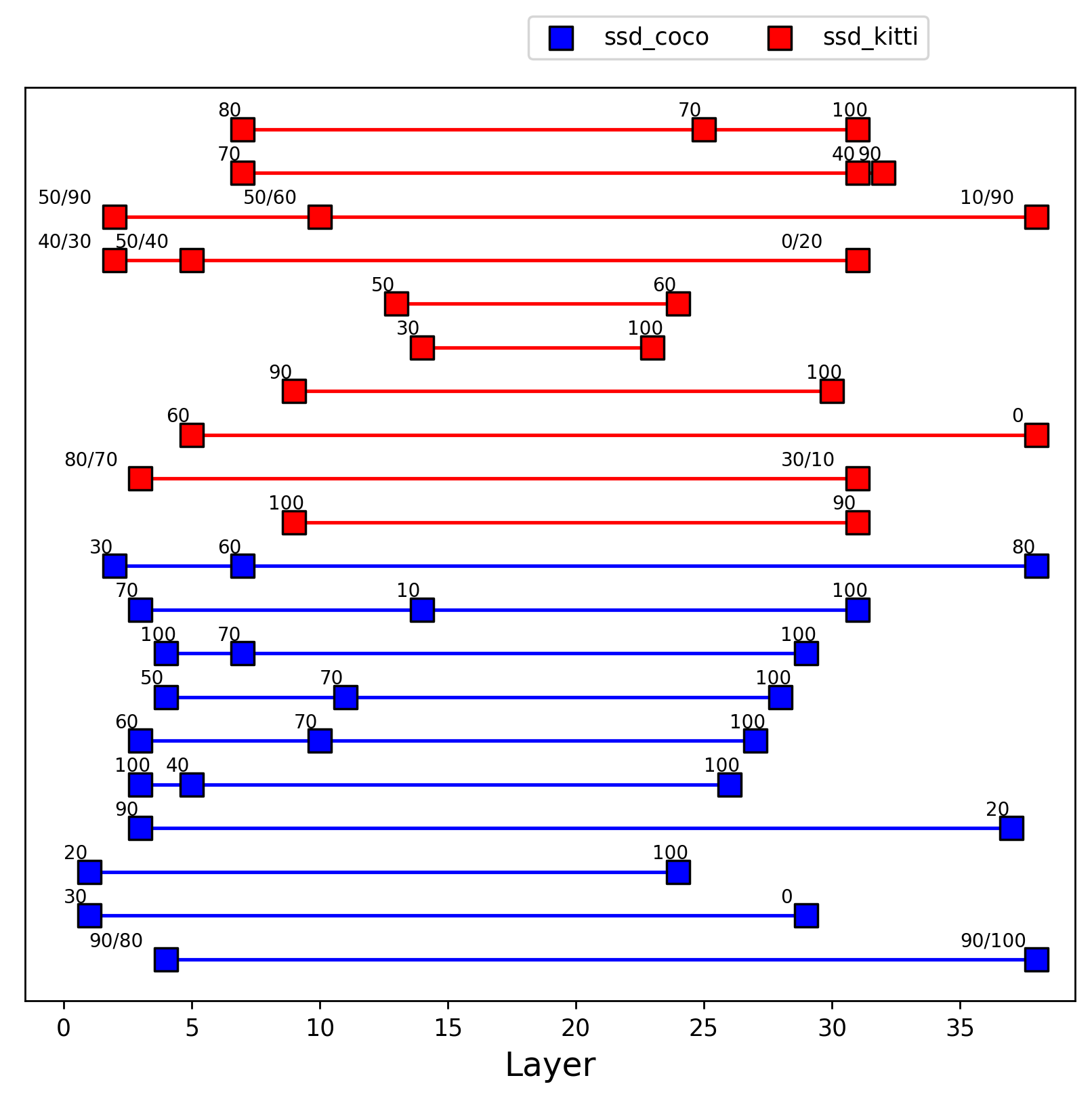}
	\end{subfigure}\\
	\begin{subfigure}{0.45\textwidth}
	(c) RetinaNet\\
	\includegraphics[width=\textwidth]{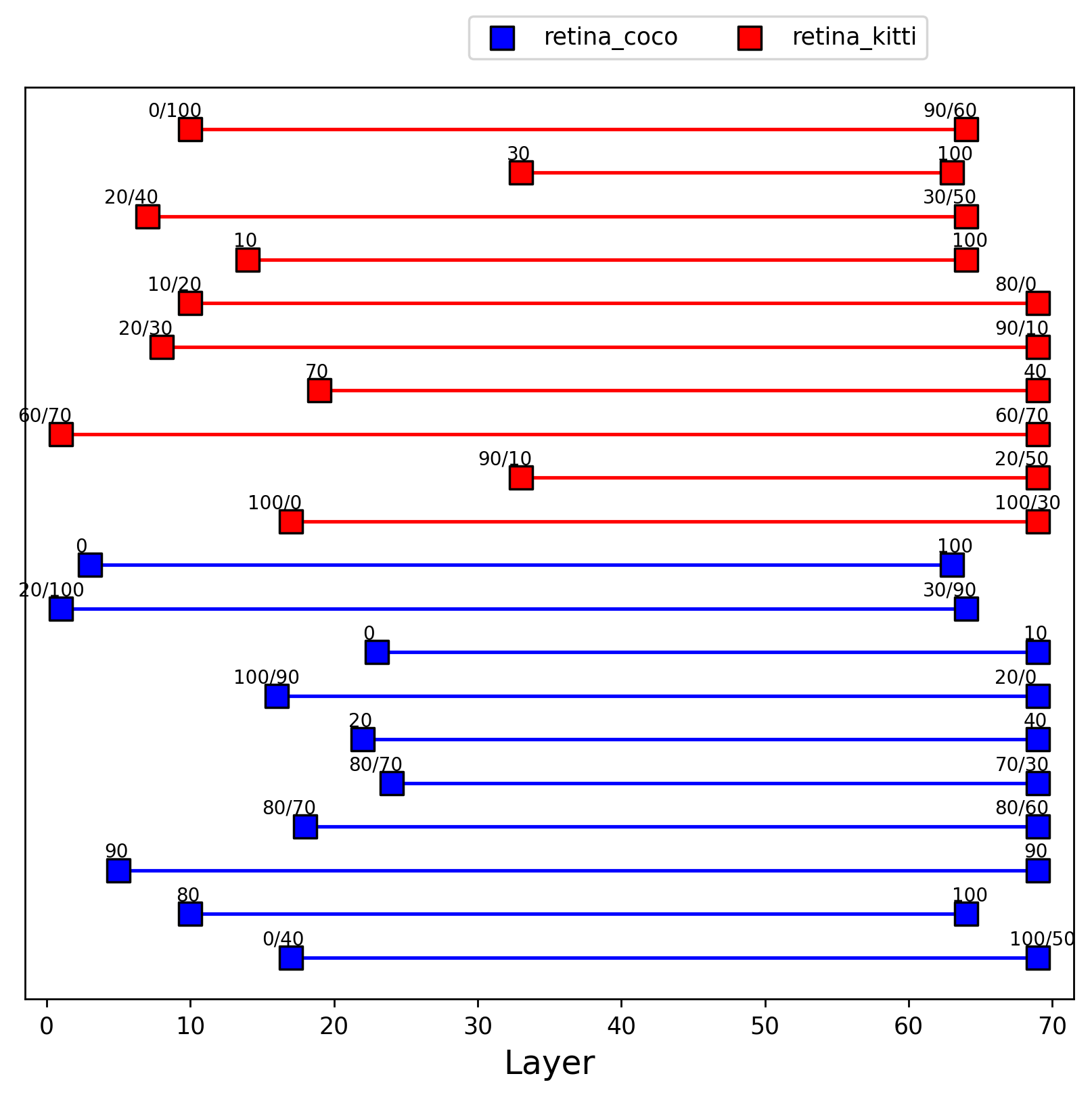}
	\end{subfigure}
	\begin{subfigure}{0.45\textwidth}
	(d) ResNet\\
	\includegraphics[width=\textwidth]{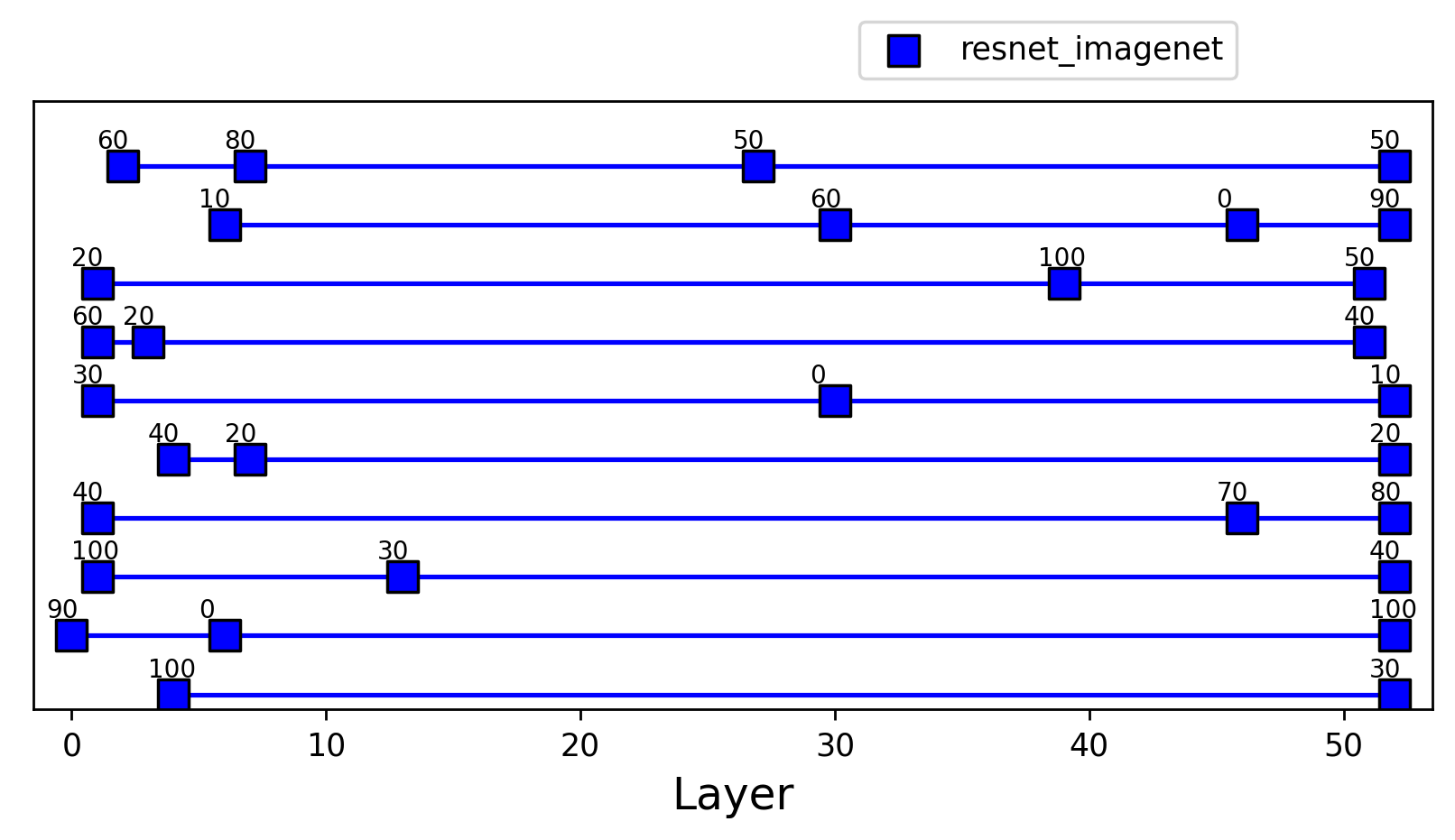}\\
	(e) AlexNet\\
	\includegraphics[width=\textwidth]{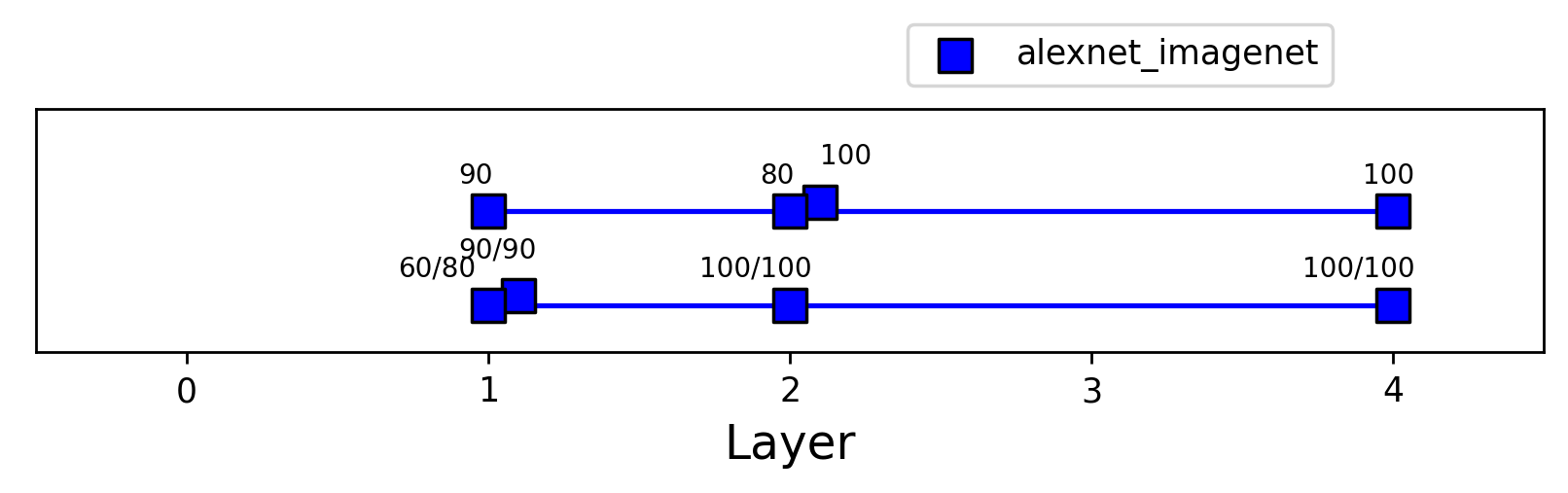}
	\end{subfigure}\\
\setlength{\belowcaptionskip}{-10pt}
\caption{Minimal combinations of features as identified by the search process in Sec.~\ref{sec:minmodel}. All combinations in (a)-(e) constitute a reduced classifier model with at least $95\%$ of the performance of the respective full model. Inset numbers designate the percentile numbers (or combinations thereof if multiple combinations are equally valid).}
\label{fig:ftcombos}
\end{figure}

\textbf{Universal Trends:}
The identified minimal feature combinations are shown in Fig.~\ref{fig:ftcombos}. We find that just $2$ features from $2$ different layers are sufficient to constitute an efficient error detector for all studied models except for AlexNet ($4$ features from $3$ layers).

Almost universally, one of the monitored layers needs to be among the very last layers of the deep neural network. Since memory faults are injected randomly across the network, errors in the last layers would go unnoticed otherwise. Only for SSD models, it turns out that most of the SDC faults occur in earlier layers, so that a supervision of the last layers is less crucial to achieve a similar statistical detection performance.
We observe that it is favorable to supervise a higher percentile (e.g., $q_{100}$) in the later layers, especially in more shallow networks (AlexNet and SSD). 
This is because in shallow networks, peak shifts have a shorter propagation path and hence it is more important to intercept faults directly. This can only be achieved by the highest percentiles. In models with ReLU activation functions (all except Yolo here), the minimum quantile  does not serve as a meaningful peak shift monitor as negative activations are clipped automatically.

A second monitoring marker should to be set in the first half of the network layer stack. This helps to identify input faults (which are interceptable from the very first layer) and discriminate them from memory faults. Either a low or high percentile can be chosen for supervision.

\textbf{Explainability:}
Given the above generalizable trends and the fully transparent nature of the classifier, we can make statements about the inner workings of the DNN that correlate a given input with an anomalous or regular outcome. 
Those statements can be interpreted intuitively by a human as a proxy of a decision, and hence qualify as an explanation \cite{BarredoArrieta2019}.

\subsection{Overhead}
\begin{figure}[htbp]
\centering
\includegraphics[width=.8\textwidth]{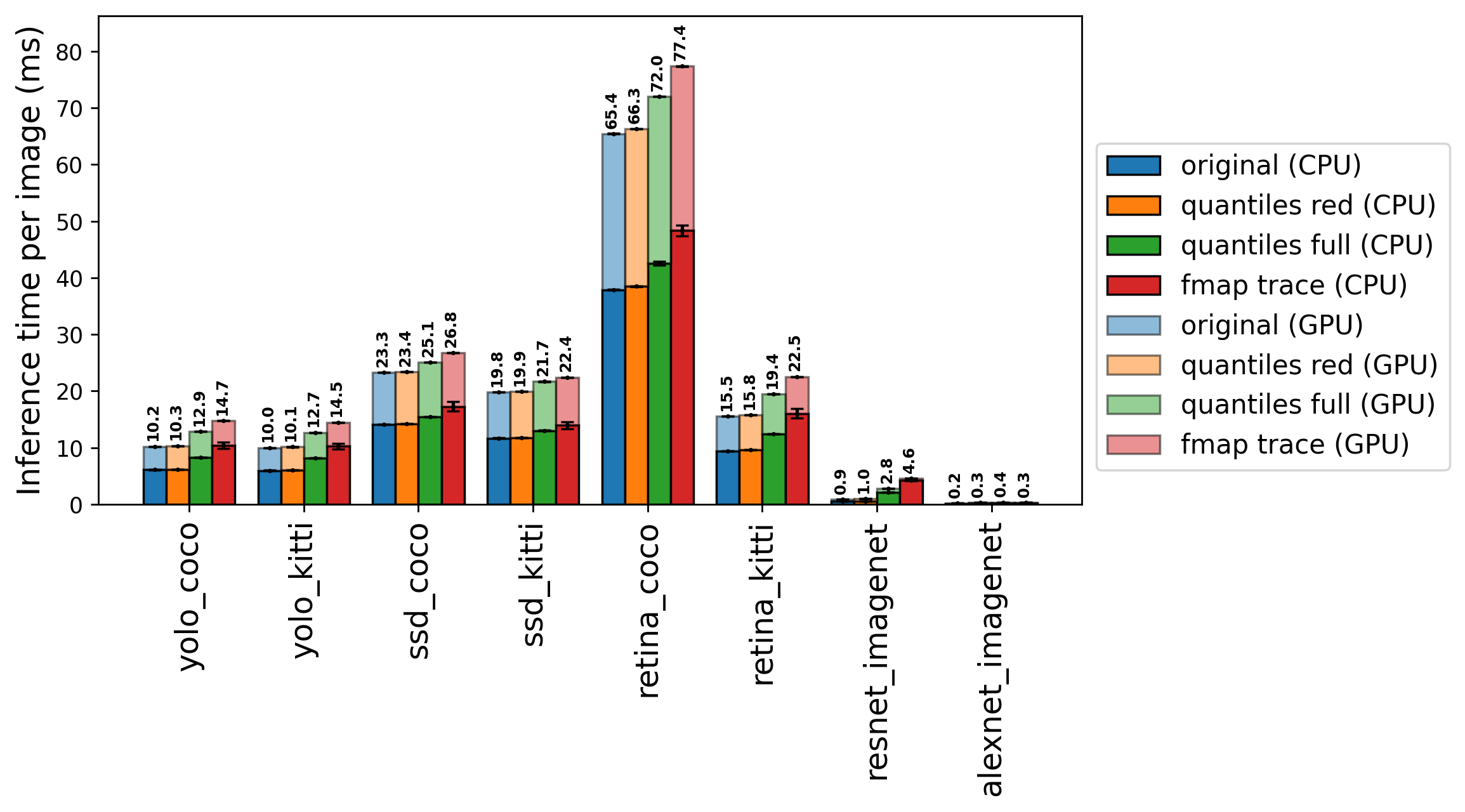}
\setlength{\belowcaptionskip}{-10pt}
\caption{Average inference time per image accumulated over CPU and GPU. We compare the original inference, reduced and full quantile monitoring, and feature map tracing (method of \cite{Schorn2018, Schorn2020}). In the setup, we run $100$ random images with a batch size of $10$ (with GPU enabled) and repeat $100$ independent runs. System specifications: Intel{\textregistered} Core\textsuperscript{TM} i9-12900K, Nvidia GeForce RTX 3090.}
\label{fig:latency}
\end{figure}

We measure the average inference time per image when running the supervised model on random input, using the \textit{Torch profiler} \cite{Paszke2019}. 
The profiled overall self compute time in Fig.~\ref{fig:latency} is shared between CPU and GPU.
Compared to the feature map tracing method of Schorn et al.~\cite{Schorn2018, Schorn2020}, the quantile operation introduces additional compute, but at the same time saves the time of storing large tensors, due to the compression of many feature sums into only a few quantiles.
Between these two opposing trends, full quantile monitoring turns out to be \textit{faster} than feature map tracing for all the studied models except for the shallow AlexNet, as shown in Fig.~\ref{fig:latency}. 
If only selected layers are monitored to create a reduced classifier, the overhead can be decreased significantly.
We find that the impact of minimal quantile monitoring on the overall inference time is between $+0.3\%$ and $+1.6\%$ for all studied object detection DNNs. For the image classification networks, on the other hand, quantile monitoring imposes a more significant overhead of $+10.7\%$ (ResNet) and $+53.8\%$ (AlexNet). This is because those networks have a much smaller number of parameters such that the relative impact of quantile extraction with respect to the total number of operations is higher. Across all models, minimal quantile monitoring is $>10\%$ faster than feature map tracing.
In absolute numbers, the respective saving in inference time can be up to ${\sim}10\text{ms}$, which is a significant improvement for applications operating at real-time, for example object detection in a self-driving vehicle.

\subsection{Comparison with Other Detector Approaches}
\label{sec:compareLR}
Alternative to a decision tree, we can deploy a linear machine learning model for error detection (similar to \cite{Schorn2020}). We study the feasibility of doing so in this section. 
For this setup, we select Yolo+Kitti to train a classifier for $1000$ epochs using the Adam optimizer and cross entropy loss. A batch size of $100$ and learning rates optimized between $1\times 10^{-4}$ and $5\times 10^{-3}$ were chosen.
In the simplest form, with a multi-layer-perceptron, the algorithmic transparency is preserved and we find $\Pcls = 86.0\%$ and $\Rcls=95.7\%$.
If more hidden linear layers are added, higher detection rates can be achieved at the cost of explainability.
For example, including one extra hidden layer with $64$ neurons \cite{Schorn2020}, we find a performance of $\Pcls = 88.9\%$ and $\Rcls=96.3\%$, with three such extra layers we obtain $\Pcls = 91.7\%$ and $\Rcls=95.1\%$.
Compared to decision trees, however, this strategy suffers from more complex hyperparameter tuning and large training times. Therefore, decision trees are a better fit for our use case.

\section{Summary and Future Work}
\label{sec:summary}
In this paper, we show that critical silent data corruptions in computer vision DNNs (originating either from hardware memory faults or input corruptions) can be efficiently detected by monitoring the quantile shifts of the activation distributions in specific layers. In most studied cases, it is sufficient to supervise two layers with one quantile marker each to achieve high error detection rates up to ${\sim}96\%$ precision and ${\sim}98\%$ recall.
We also show that the strategic monitoring location can be associated with the concept of intercepting bulk and peak activation shifts, which gives a {\em novel, unifying perspective on the dependability of DNNs}.
Due to the large degree of information compression in this approach, the compute overhead of the approach is in most cases only between $0.3\%$ and $1.6\%$ compared to the original inference time, and outperforms the comparable state of the art.
In addition, we show that the method contributes to the model's explainability as the error detection decision is interpretable and transparent.
For future work, we can further guide the search for optimized minimal feature combinations, for example, by taking into account specifics of the model architecture. \\

\textbf{Acknowledgement:}
We thank Neslihan Kose Cihangir and Yang Peng for helpful discussions.
This project has received funding from the European Union's Horizon $2020$ research and innovation programme under grant agreement No $956123$.
This work was partially funded by the Federal Ministry for Economic Affairs and Climate Action of Germany, as part of the research project SafeWahr (Grant Number: $19A21026C$), and the Natural Sciences and Engineering Research Council of Canada (NSERC).

\bibliographystyle{splncs04}
\bibliography{ActBib}

\end{document}